\documentclass[journal]{IEEEtai}

\usepackage{footnote}
\usepackage[colorlinks,urlcolor=blue,linkcolor=blue,citecolor=blue]{hyperref}
\usepackage{color}
\usepackage{filecontents}
\usepackage{amssymb,amsmath,amsthm, empheq}
\usepackage{mathtools} 
\usepackage{verbatim}
\usepackage[margin=1in]{geometry}
\usepackage[utf8]{inputenc}
\usepackage{csquotes}
\usepackage[linesnumbered,ruled,lined,boxed]{algorithm2e}
\usepackage{dsfont}
\usepackage{setspace}
\usepackage{comment}
\usepackage[english]{babel}
\usepackage{array}
\usepackage{float}
\usepackage{multirow}
\usepackage{graphicx}
\usepackage{caption}
\usepackage{subcaption}
\graphicspath{ {./imgs/} }

\newcommand{\norm}[1]{\|#1\|}
\newtheorem{theorem}{Theorem}

\newtheorem{definition}{Definition}
\newtheorem{proposition}{Proposition}

\begin{document}

\title{On Tuning Neural ODE for Stability, Consistency and Faster Convergence}
\author{Sheikh Waqas Akhtar
		\thanks{S. W. Akhtar is with the University of Central Punjab, Lahore, Pakistan (e-mail: sheikh.waqas@ucp.edu.pk).}}	
\markboth{Journal of IEEE Transactions on Artificial Intelligence, Vol. 00, No. 0, Month 2020}
{Akhtar \MakeLowercase{\textit{et al.}}: On Tuning Neural ODE}

\maketitle	
\begin{abstract}
	Neural-ODE parameterize a differential equation using continuous depth neural network and solve it using numerical ODE-integrator. These models offer a constant memory cost compared to models with discrete sequence of hidden layers in which memory cost increases linearly with the number of layers. In addition to memory efficiency, other benefits of neural-ode include adaptability of evaluation approach to input, and flexibility to choose numerical precision or fast training. However, despite having all these benefits, it still has some limitations. We identify the ODE-integrator (also called ODE-solver) as the weakest link in the chain as it may have stability, consistency and convergence (CCS) issues and may suffer from slower convergence or may not converge at all. We propose a first-order Nesterov's accelerated gradient (NAG) based ODE-solver which is proven to be tuned vis-a-vis CCS conditions. We empirically demonstrate the efficacy of our approach by training faster, while achieving better or comparable performance against neural-ode employing other fixed-step explicit ODE-solvers as well discrete depth models such as ResNet in three different tasks including supervised classification, density estimation, and time-series modelling.    
\end{abstract}

\begin{IEEEImpStatement}
	The broader impact of this work, if any, would be an improvement in the modeling tools for machine learning tasks like regression, classification, generative modeling etc. This work is an effort towards making machine learning algorithms faster, stable and consistent. We cannot speculate about or foresee any negative impact or mis-use of this work.      
\end{IEEEImpStatement}


\section{Introduction}

\IEEEPARstart{R}{esidual} networks or ResNets models data by learning the dynamics of data governed by an ordinary differential equation discretized in time. This discretization is represented by the number of layers (depth) of neural network. 
\begin{equation}\label{eq1}
\textbf{h}_{t+1} = \textbf{h}_{t} + \mathit{f}(\textbf{h}_{t}, \theta_{t})
\end{equation}
The continuous time counterpart of \ref{eq1} is  
\begin{equation}\label{eq2}
\frac{d\textbf{h}(t)}{dt} = \mathit{f}(\textbf{h}(t), t, \theta)
\end{equation}
Neural-ODEs parameterize \ref{eq2} using a neural network and solve it using a numerical ode-solver which approximately integrates the dynamics within the desired tolerance in error. $\textbf{h}(0)$ is the input layer and  $\textbf{h}(T)$ is the output layer  producing the solution of ODE at time $T$. This model was proposed by \cite{Chen2018}. They tested neural-ODE for a variety of machine learning tasks and showed that its performance is quite competitive to a deep residual network. The main advantage of neural-ODE over traditional ResNet is its memory efficiency while performance is comparable (See Table 1 \cite{Chen2018}). Neural-ODE used adjoint sensitivity method \cite{Pont1962} to compute gradients of loss function with respect to weights of ODE-network. The adjoint sensitivity method trains the model with constant memory cost as a function of depth, which is main advantage over discrete depth models such as ResNet. However, in terms of time complexity, it is not much better than ResNet and in some cases, its even worse than ResNet. The main reason is the usage of numerical ODE-solver which is effectively a black-box module in the neural-ode. ODE-solver requires a number of forward evaluations (NFE) of hidden state dynamics to produce the output. The number of forward evaluations refers to time discretizations required to achieve the desired error tolerance, set by the practitioner. How many NFEs will it take to bring the error down to tolerence threshold, is solver dependent. The practitioner does not have any control on it. Thus, convergence can be too slow in some instances.   \\ 
In addition to this, the design of ode-solver can pose stability and consistency issues as well and the solution may not even converge at all. Therefore, it is essential to choose an ode-solver that is free from such design problems or if the practitioner is designing an ode-solver, the design should be constrained to satisfy consistency, stability and convergence conditions (see Appendix- \ref{appendix-A} for details). \par
In this paper, we investigate the question of how to enforce the stability, consistency and covergence conditions on an ODE solver used in neural-ode. We closely follow the findings of \cite{Sci2017} who studied the relationship between numerical ODE solvers and gradient based optimization algorithms. They established that the linear multi-step ode solvers under the constraints of stability, consistency and convergence, can be modeled as gradient based optimizers. Building upon these findings, we propose a novel neural ode architecture with a nesterov accelerated gradient (NAG) based ode solver tuned for CCS conditions. We compared its performance with neural-odes employing various explicit numerical ode solvers, as well as with discrete depth counterpart, ResNet. 

\section{Related Work}
There have been some efforts to make neural odes more stable and converge faster. Most notably, \cite{Finlay2020} \cite{Kelly2020} and \cite{Ghosh2020} proposed a regularization based approach to train neural ode in which the regularization terms were specifically designed to stabilize the dynamics of model. This has an effect on convergence rate as well. The model learns a smoothed out dynamics faster as compared to a coarse one but this increases the error. So, regularization based approach requires us to make a trade-off between speed and performance.
\cite{Dup2019} proposed a method to increase the representation power of the dynamics by uplifting it to higher dimensions. To do so they augmented the dynamics features space with additional empty dimensions and showed that by doing this their model learned a simplified dynamics which needed much less NFEs to learn it as compared to non-augmented couterpart in which NFEs grow exponentially during training. Other works used data-control\cite{Stefano2020} and depth-variance \cite{Stefano2020},\cite{Tan2022}.
\par
Another branch of research involves learning the higher-order dynamics using Nerual-ODEs.
These models take advantage of the learned acceleration (a 2nd order dynamics) thereby reducing the NFEs in solving both forward and backward calls and speed-up the learning. \cite{Norcliffe2020} learns to solve a second ODE as a system of first-order ODEs. \cite{Xia2021} also solves a second-order ODE but has a constant momentum factor to speed-up the learning. \cite{Nguyen2022} learns a seconder-order ODE limit of the Nesterov accelerated gradient (NAG) with a time-dependent momentum factor. Their approach, although bears resemblence with our method in using NAG but it is an entirely different breed and focusses on learning the acceleration of the dynamics. Their method has shown improvement in speed, performance and stability over Neural-ODE learning first-order dynamics. Their notion of stability is limited to the choice of step-size, showing that the performance remains unaffected on changing the step-size of ODE-solver. Contrary to this, we have focussed on the ODE-solver itself and constrained it be zero-stable, consistent and convergent and learn first-order dynamics using it.  

\section{Background}
\subsection{Initial Value Problem}
The dynamics or flow of the state vector $x(t)$ of a dynamical system can be modeled by an ODE $\frac{dx(t)}{dt} = f(x(t), t, \theta)$. Given an initial state $x(t_{0})$, the state at a later time $x_{t}$ is given by:
\begin{equation}\label{eq3}
x(t_{1}) = x(t_{0}) + \int_{t_{0}}^{t_{1}}f(x(t), t, \theta)dt 
\end{equation}
is called an initial value problem (IVP) whereas $f$ represents the dynamics or evolution of state vector $x(t)$ with respect to time. For example, $f$ could describe the equation of motion of a particle, transmission rate of a virus across a population. 

\subsection{Neural ODE}
Neural-ODE is a neural network architecture which is continuous depth analogue of ResNet \cite{He2016}. Lets recall how ResNet solves the IVP whose dynamics are not known. ResNet has multiple residual blocks. Each succeeding block represents a discretization of the dynamics in time. A residual network with N blocks will produce an output $t_{N}$ steps forward in time. In a neural-ODE, the dynamics $\dot x$ can be approximated by a moderately sized neural network, its parameters $\theta$ trained by an optimization algorithm e.g gradient descent. The output of dynamics network is passed to a numerical ode-solver which integrates it upto the specified time $T$. The output at time $T$ is the solution of IVP after $T$ time step.\\
However residual neural network has a large memory footprint and suffers from the problem of diminishing and exploding gradients. \cite{Chen2018} proposed a neural-ODE trained using adjoint sensitivity method \cite{Pont1962} which substantially reduced its memory footprint but this method used adaptive step-size numerical ODE solver which acted as a black-box inhibiting control over the number forward steps required by the solver. As a result, neural ode method is often slower than fixed-depth residual neural network. Moreover, it also has stability and consistency issues (see Experiments section for more details).

\section{Consistent, Convergent and Stable ODE-Solver}
Generally, the integral in \ref{eq3} has no closed form analytic solution and must be approximated numerically using an ode-solver which integrates the dynamics on a finite interval  $[0, t_{max}]$. The time discretization, also called step-size $h_{k} = t_{k} - t_{k-1}$ is assumed constant for the sake of simplicity. Our aim is to minimize the approximation error $\|x_{k}-x(t_{k})\|$ for $k \in [0,t_{max}/k]$. $x_{k}$ is the predicted value and $x(t_{k})$ is the true value. 
\\ ODE-solver has a very crucial role in neural-ode and its design parameters should not be such that it has stability, consistency and convergence issue. Therefore, we need to tune our ode-solver for CCS condition. We take linear multi-step method as a case study to tune it for CCS conditions. The reason for its selection is that a large number of off-the-shelf ode-solver such as Euler method, Adam-Bashforth method, Adam-Moulton methods, and the backward differentiation formula (BDFs) belong to and are special cases of linear multi-step methods.   \par
\subsection{Linear Multi-step Methods}
Linear multi-step method is an auto-regressive method and uses several past iterates to predict the next value. It is given by 
\begin{equation}\label{lms}
x_{k+s} = - \sum_{i=0}^{s-1} a_{i}x_{k+i} + h \sum_{i=0}^{s}b_{i}g(x_{k+i}), \hspace{5mm} k \ge 0, 
\end{equation}
where $a_{i}, b_{i} \in \mathrm{R}$ are the parameters of multi-step method and s represents the number of past values required. Each new value $x_{k+s}$ is a function of the information given by the s previous values. If $b_{s}=0$, each new value is given explicitly by the s previous values. Such a method is called an explicit method. Otherwise, new value not only depends on past s values but also on some function $g$ of new value. This requires solving a nonlinear (in general) system of equations at each step. Such a method is called implicit method. \\
Let's now present an alternate notation for \ref{lms}. We define the first and second characteristic polynomials of \ref{lms} by
\begin{equation}
P(\zeta) := \sum_{i=0}^{s} a_{i}\zeta^{i}, \hskip 3em  Q(\zeta) := \sum_{i=0}^{s}b_{i}\zeta^{i}
\end{equation}
where $\zeta \in \mathcal{C}$ is a dummy variable. \ref{lms} can now be written in the form:
\begin{equation}
P(E)x_{k} = hQ(E)g_{k}, \hspace{5mm} for \ every \ k \ge 0
\end{equation}
where $E$ is the forward shift operator which maps $Ex_{k} \rightarrow x_{k+1}$, $P$ and $Q$ are polynomials of degree s with coefficients $a_{i}$ and $b_{i}$ respectively. P is also monic i.e $a_{s} = 1$ and h is the step-size. \\

\subsection{Tuning linear multi-step method with CCS conditions}
Scier et.al \cite{Sci2017}, showed that tuning a 2-step linear ode-solver with CCS conditions (see Appendix \ref{appendix-A} for details) can be posed as a constrained optimization problem ensuring that the constraints on the coefficients  \ref{ccs-cond}, of characteristic polynomials are satisfied. Parameters of a tuned 2-step linear method, called $\mathcal{M}$, are

\begin{equation}\label{M}
\mathcal{M} = 
\begin{aligned}
\begin{cases}
a(z) &= \beta - (1 + \beta)z + z^{2}, \\
b(z) &= -\beta (1-\beta) + (1-\beta^{2})z, \\
h &= \frac{1}{L(1-\beta)}
\end{cases} 
\end{aligned}
\end{equation}
where $\beta$ is a scalar and depends on Lipschitz constant L of the dynamics.

\section{Nesterov's accelerated gradient based optimizer as an ODE Solver}
We show here that a tuned linear two-step method $\mathcal{M}$ can be posed as Nesterov's accelerated gradient (NAG) method. NAG is a first-order optimization algorithm with a "corrected momentum". Momentum based methods accelerate learning by adding a momentum term in the gradient descent update rule. This momentum term is the accumulated gradients from previous iterations. This allows the optimization algorithm to avoid getting stuck in a local minima. Standard momentum based method such as Polyak Heavy Ball method \cite{Polyak1964} compute gradient at current iteration, add momentum term and take a jump in the direction of this update. However, if that new position is not a good one, algorithm will have to improve its results again and this will make the algorithm too slow. Nesterov's method provides a remedy by correcting the momentum term. It first makes an interim update by jump in the direction of accumulated gradient, and if it is a bad position, then it will take a corrective measure and direct the update back towards the current position. It can be described by two sequences $x_{k}$ and $y_{k}$. $y_{k}$ is the interim update
\begin{align}
y_{k+1} &= x_{k}  - \frac{1}{L}\nabla f(x_{k}) \label{nesterov:eq1}\\   
x_{k+1} &= y_{k+1} + \beta(y_{k+1} - y_{k}) \label{nesterov:eq2}	
\end{align}
After some basic algebra, update equation \ref{nesterov:eq2} can written without interim update terms $y_{k+1}$ and $y_{k}$ as
$$\beta x_{k} -(1 + \beta)x_{k+1} + x_{k+2} = \frac{1}{L}(-\beta(-\nabla f(x_{k})) + (1+\beta)(-\nabla f(x_{k+1}))) $$
Consistency of the method is then ensured by checking
\begin{align*}
a(1) &= 0 \qquad \text{Always Satisfied} \\
\acute{a}(1) &= b(1) \quad \implies \ h = \frac{1}{L(1-\beta)} 
\end{align*}
After collecting the parameters of polynomials $\rho(z)$ and $\sigma(z)$, we see that it is indeed equal to $\mathcal{M}$ \ref{M}.
We propose to use nesterov gradient descent as an ode-solver. Pseudo-code of Nesterov ODE-Solver \ref{algo1} is outline below:

\begin{algorithm}
	\caption{Nesterov ODE Solver}
	\label{algo1}
	\KwIn{$f, x_{k}, y_{k}, L, \beta$} 
	\KwOut{$x_{k+1}, y_{k+1}$}
	$y_{k+1} = x_{k} - \frac{1}{L}\nabla f(x_{k})$ \tcp{Interim new value}
	$x_{k+1} = y_{k+1} + \beta (y_{k+1} - y_{k})$ \tcp{New value}
	\textbf{return} $x_{k+1} , y_{k+1}$
	
\end{algorithm}

\section{A Discussion on the relationship between Resnet, RNN and Neural-ODE}
Resent, Neural-ODE and RNN are closely connected family of neural networks. Neural-ODE was posed as a continuous time variant of Resnet with significant reduction in memory footprint, with comparable or somewhat better performance than ResNet \cite{Chen2018}. The close relation of Resnet and RNN has been studied in \cite{Poggio2016} who showed that a shallow RNN and a very deep ResNet with a weight sharing among the layers are equivalent and have similar performance. RNN and Neural-ODE are also related. \cite{Adeline2021} studied relationship between RNN and Neural-ODE and showed that a kernelized RNN can be interpreted as Neural-ODE. Our work also sheds lights onto this interesting relationship and shows that both RNN and Neural-ODE show similar advantage vis-a-vis ResNet, thus suggesting their close connection empirically. Using optimization algorithm allowed us to open the black-box of ODE-solver and discover nuanced similarities and differences between neural-ode and RNN.         

\subsection{RNN and Neural-ODE- Two faces of the same coin}
\begin{figure*}
	\centering
	\includegraphics[width=40pc]{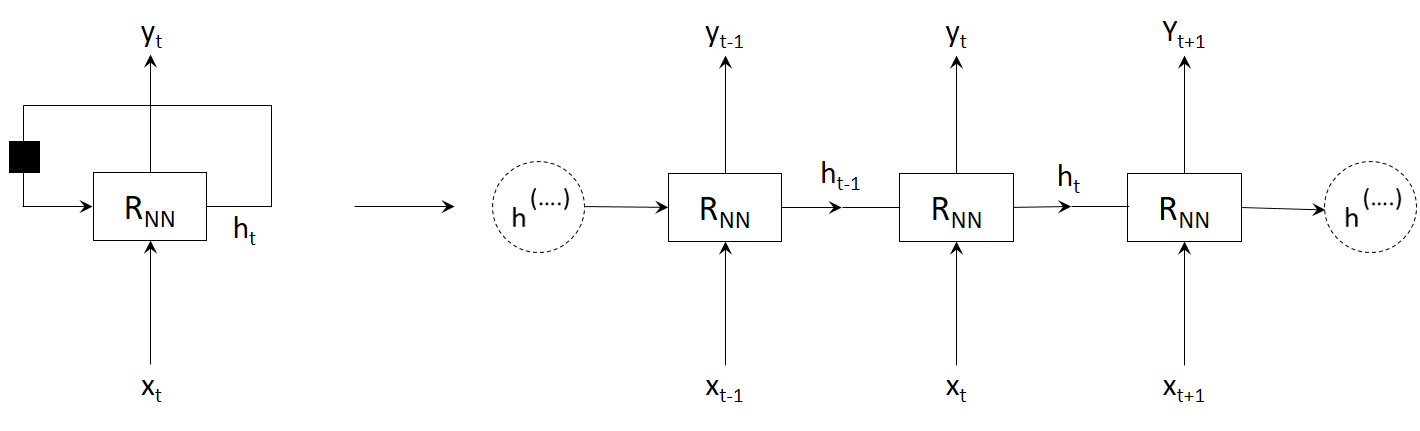}
	\caption{Unrolled RNN}
	\label{fig:rnn}
\end{figure*}

\begin{figure*}
	\centering
	\includegraphics [width= 40pc]{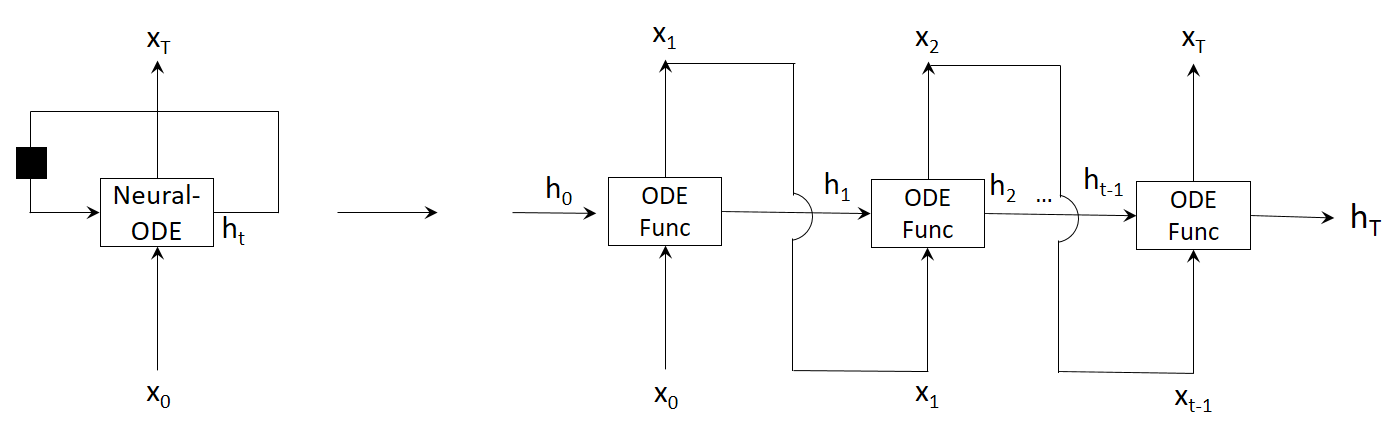}
	\caption{Unrolled Neural-ODE}
	\label{fig:neural-ode}
\end{figure*}

RNN and Neural-ODE are closely related to each other however, there are some differences as well. In RNN, hidden state and output are functions of learned weights. Weight matrices of both hidden state and output are different. Although there is weight sharing in layers. In Neural-ODE, both hidden state and output are a function of dynamics described by the weight matrix of ode-network. This means that in neural-ode there is weight sharing in hidden state and output across all time stamps. The update equations of hidden state and output in neural-ode is dependent on the  ode-solver. For example, if Nesterov ODE-solver is being used, the interim update equation \ref{nesterov:eq1}  will be the hidden state and corrected update equation \ref{nesterov:eq2} will be the output. Input to the ODE-Solver at time $t$ is the output obtained at the previous time step $t-1$. This is similar to one-to-many architecture of RNN. See Figure \ref{fig:rnn} and Figure \ref{fig:neural-ode}  

\section{Experiments}
We experimented on toy and real data. In toy experiments, we solved an ODE using neural-ode employing different ode-solvers and compared their performance. The caveat is that these solvers suffer from instability, inconsistency and/or solution divergence issues. These experiments validate the necessity of using tuned ode-solvers in Neural-ODEs. For experiments on real data, we considered three different learning tasks for empirical evaluations: supervised classification, modeling of time series \cite{Ruba2019} and density estimation.

\subsection{Experiments on Toy data}
We used a 1-layered MLP (with 50 neurons in the hidden layer and tanh activation) to model the differential equation (an IVP). The input and outputs are 2-dimensional. The output of MLP is fed to the ode-solver (described in the following examples) to solve the initial value problem, on a range of step sizes. The initial value problem to be solved is
\begin{equation}\label{toy_ivp}
\begin{aligned}
y^{\prime} &= f(x,y), \quad  y(0) = [0.5, -3]^{T}, \quad  x \in [0,1]\\
y &= [u,v]^{T},\\
f(x,y) &= [v, v(v-1)/u]^{T}
\end{aligned} 
\end{equation}
It can be verified (using Theorem 1.1 of \cite{Lambert1991}) that \ref{toy_ivp} has a unique solution. The unique exact solution is 
\begin{equation*}
u(x) = (1+3exp(-8x))/8, \quad v(x) = -3exp(-8x)
\end{equation*} 
The solution decays in the sense that both $|u(x)|$ and $|v(x)|$ decrease monotonically as x increases from 0 to 1.

\noindent
\textbf{Example 1:}
\begin{equation}\label{example1}
y_{t+2} + y_{t+1} - 2y_{t} = \frac{h}{4}[f(x_{t+2},y_{t+2}) + 8f(x_{t+1},y_{t+1}) +3f(x_{t},y_{t})]
\end{equation}
This method is consistent but zero-unstable and therefore divergent. Results in Table \ref{table:ex1} show that the solution diverges irrespective of the step-size.
\newline
\begin{table}
	\centering
	\begin{tabular}{|c|c|c|c|c|}
		\hline
		\multirow{3}{*}{MAE} & \multicolumn{4}{c|}{Step-Size (h)} \\
		\cline {2-5}
		& h = 0.1 & h = 0.01 & h=0.001 & h = 0.0001 \\
		\cline{2-5} 
		& - & - & - & - \\
		\hline
		
	\end{tabular}
	\caption{Mean absolute Error for different step-sizes h, at 2000\textsuperscript{th} training Iteration. - means that the number is too big for Python to show, so it returns NAN (not a number)}
	\label{table:ex1}
\end{table}

\noindent
\textbf{Example 2:}
\begin{equation}\label{example2}
y_{t+2} - y_{t+1} = \frac{h}{3}[3f(x_{t+1},y_{t+1}) -2f(x_{t},y_{t})]
\end{equation}
This method is zero-stable but inconsistent and therefore divergent. ODE-Solver in example 1 was divergent due to instability which led to an explosion of error. In example 2, divergence is caused by inconsistency and does not lead to an explosion but manifests itself in a persistent error which refuses to decay to zero even at very small step-sizes, as evidenced in the results shown in Table \ref{table:ex2}. 
\newline
\begin{table}
	\centering
	\begin{tabular}{|c|c|c|c|c|}
		\hline
		\multirow{3}{*}{MAE} & \multicolumn{4}{c|}{Step-Size (h)} \\
		\cline {2-5}
		& h = 0.1 & h = 0.05 & h=0.025 & h = 0.0125 \\
		\cline{2-5} 
		& 0.1296 & 0.1293 & 0.1410 & 0.1378 \\
		\hline
		
	\end{tabular}
	\caption{Mean absolute Error for different step-sizes h, at 2000\textsuperscript{th} training Iteration}
	\label{table:ex2}
\end{table}

\noindent
\textbf{Example 3}
\begin{multline}\label{example3}
y_{t+3} + \frac{1}{4}y_{t+2} - \frac{1}{2}y_{t+1} - \frac{3}{4}y_{t} = \frac{h}{8}[19f(x_{t+2},y_{t+2}) \\
 + 5f(x_{t}, y_{t})]
\end{multline} 
This method is consistent and zero-stable and therefore convergent. The variation in error by changing step-size show that there exist some optimal value of h for which the ode-solver performs best. See results in Table \ref{table:ex3} 
\newline

\begin{table}
	\centering
	\begin{tabular}{|c|c|c|c|c|}
		\hline
		\multirow{3}{*}{MAE} & \multicolumn{4}{c|}{Step-Size (h)} \\
		\cline {2-5}
		& h = 0.1 & h = 0.05 & h=0.025 & h = 0.0125 \\
		\cline{2-5} 
		& 0.1834 & 0.0632 & 0.0720 & 0.0470 \\
		\hline
		
	\end{tabular}
	\caption{Mean absolute Error for different step-sizes h, at 2000\textsuperscript{th} training Iteration}
	\label{table:ex3}
\end{table}

\subsection{Experiments on Real data}
\subsubsection{Supervised Learning}
We used MNIST dataset for classification using two types of neural networks. We used the same architecture as in \cite{Chen2018}\footnote[1]{https://github.com/rtqichen/torchdiffeq}. Training details are discussed in Appendix.
\begin{itemize}
	
	\item A Residual network with twice input downsampling, followed by six standard residual blocks \cite{He2016}.
	\item A Neural-ODE, in which the residual blocks are replaced by an ODESolve module which incorporates a numerical ODE-Solver\cite{Chen2018} . 
\end{itemize}

We trained both Neural-ode and Resnet using SGD for classification on MNIST dataset. For Neural-ODE, we used fixed-step explicit solvers such as Euler, Nesterov (proposed), AdamsBashforth with order 4 and Runge-Kutta4 with order 5 (also known as dopri5 \cite{Dormand1980}). Nesterov uses accelerated nesterov gradient descent method as ODE-solver. Gradients were computed using Pytorch's Autograd and Adjoint Method\cite{Pont1962}. Training algorithm of Neural-ODE is outline in \ref{algo2}:

\begin{algorithm}
	\caption{Training a Neural-ODE for Supervised Learning}
	\label{algo2}
	\KwIn{, Training dataset}  
	\KwOut{Learned weights $\theta$}  
	\For{$batch \leftarrow 1$ \KwTo $N$}{
		
		$Y_{pred}$ = ODESolver$(f_{\theta})$, Batch-$y_{0}$, Batch-t) \tcp{$f_{\theta}$ - a NN based approximation of ODE-function}
		loss = $\frac{1}{K} (Y_{pred} - Y_{true})^{2}$  \tcp{K samples in each batch}
		$\frac{\partial L}{\partial \theta}$ = GradientComputation(loss) \tcp{Autograd or Adjoint Method}	
		$ \theta = \theta - lr * \frac{\partial L}{\partial \theta}$ \tcp{weights update}
	}
	return $\theta$
\end{algorithm}     

\noindent
\textbf{Performance of ODE-Solvers} \\
Result in Table \ref{perf_table} empirically prove the computational efficiency of Nesterov ODE-solver over Resnet and Neural-ODEs using explicit ODE-solvers for classification task using Autograd and Adjoint method for gradient computation. In terms of performance, it is better or at least comparable with other techniques. Despite providing significant improvement in space complexity, Neural-ODE had worse time complexity than ResNet, in practice. These result show that we have not only overcome that drawback but also achieved better performance than ResNet.  

\begin{table*}[htbp]
	
	\begin{tabular}	{|c|c|m{2.5cm}|c|c|c|}
		\hline
		Network & ODE-Solver & Gradient Computation & Validation Acc & Test Acc & Time(sec) \\
		\hline
		\multirow{8}{*}{Neural-ODE} & \multirow{2}{*}{Euler} & Autograd & 0.9344 & 0.9871 & 718 \\
		&  & Adjoint & 0.9104 & 0.9641 & 680 \\
		\cline{2-6} 
		&\multirow{2}{*}{Nesterov} & Autograd & 0.9348 & \textbf{0.9888} & \textbf{673} \\
		&	& Adjoint & 0.9335 & 0.9826 & \textbf{678} \\
		\cline{2-6}
		& \multirow{2}{*}{dopri5} & Autograd & 0.9317 & 0.9830 & 978 \\
		&	& Adjoint & 0.9335 & 0.9851 & 1075 \\
		\cline{2-6}
		& \multirow{2}{*}{AdamBashforth} & Autograd & 0.9283 & 0.9789 & 791 \\
		&	& Adjoint & 0.9312 & \textbf{0.9862} & 824 \\
		\cline{2-6}
		\hline
		ResNet & -	& Autograd & 0.9299 & 0.9826 & 773 \\
		\hline

	\end{tabular}
	\caption{Classification Accuracy on Validation and Test set of Neural-ODE with various ODE-Solvers and ResNET. \textbf{Higher is better.} Training time in seconds (\textbf{Lower is better}).}
	\label{perf_table}
\end{table*}

\begin{figure}[h!]
	\centering
	\includegraphics[width= 20pc]{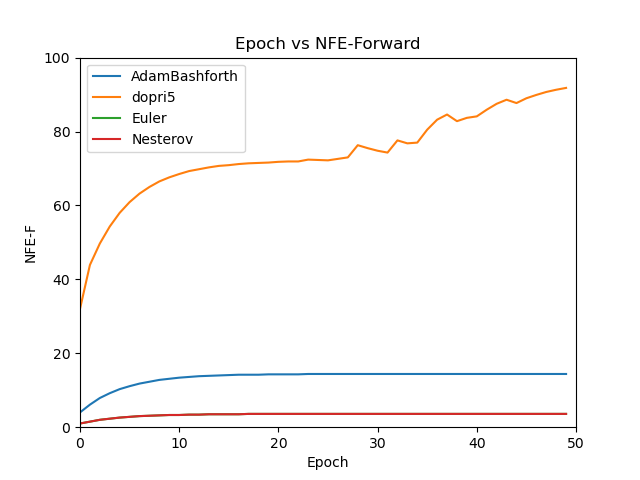}
	\caption{Training Epoch vs NFF-Forward}
	\label{nfe}
\end{figure}

\noindent
\textbf{Number of Forward Evaluations (NFE-F) of different ODE-Solvers}
Neural-ODE is a single hidden layer network. The hidden layer is called ODE-Solver and as the name implies has a numerical ODE-solver embedded in it. The concept of depth in Neural-ODE is not clearly defined. The number of forward evaluations (NFE-F) of the hidden state dynamics is analogous to depth of the neural network. NFE is determined by the ODE solver and depends on the initial state. As the model becomes increasingly complex during the training, ode-solver adapts itself to the model by increasing the number of forward evaluations. NFE-F also depends on the tolerance threshold set for the ODE-Solver. ODE solvers increase the NFE-F until the error is reduced to within the tolerance threshold. Tuning the tolerance threshold is basically making a trade-off between precision and computational cost. One could train for higher precision, but that would require more NFEs and hence has more computational cost. Results in Figure \ref{nfe} show that Euler and Nesterov have the lowest NFEs. Their NFE curves are identical, perhaps due to the fact that both of these are first-order gradient based optimizers and the only differnce is that Nesterov  applies a "momentum" to further speed-up learning. 
a lower accuracy at test time.

\begin{figure}[h!]
	\centering
	\includegraphics[width=20pc]{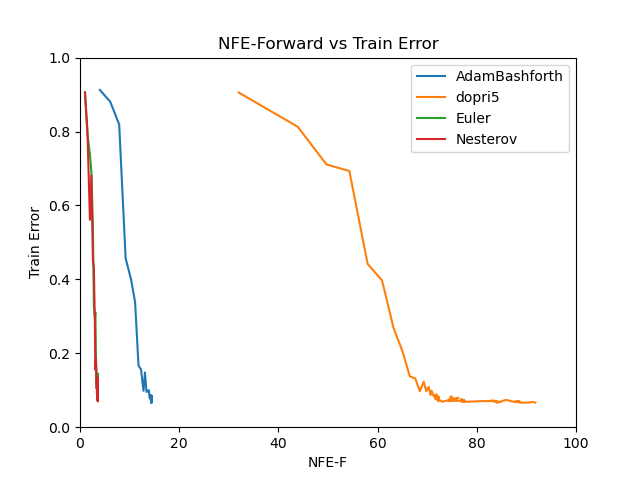}
	\caption{NFE vs. Training Error}
	\label{loss}
\end{figure}

\noindent
\textbf{Effect of NFE on loss} 
Figure \ref{loss} shows that both Euler and Nesterov gradient based solvers sharply reduce training error while keep NFE to a lower value as compared other higher-order methods like dopri5 and AdamBashforth.

\begin{figure}
	\centering
	\includegraphics[width= 20pc]{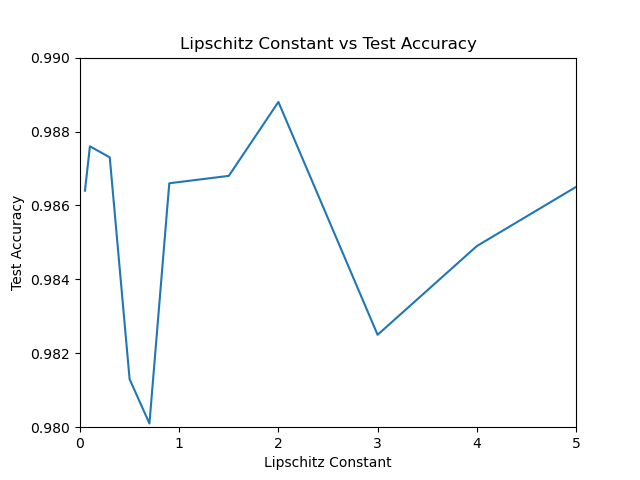}
	\caption{Lipschitz constant vs. Test Accuracy in Nesterov ODE-Solver based Neural-ODE}
	\label{lipschitz}
\end{figure}

\noindent
\textbf{Effect of Lipschitz constant in Nesterov ODE-Solver} 
Figure \ref{lipschitz} shows that Test Accuracy of the Nesterov ODE-Solver is dependent on the Lipschitz constant of dynamics. Choosing this hyperparameter poorly, may deteriorate the performance of ODE-Solver. If there is no prior knowledge about the dynamics, lipschitz constant has to be estimated using data-driven methods. An unbiased estimate of lipschitz constant is given by (Theorem 1.1 of \cite{Lambert1991}).
\begin{equation}\label{L-estimate}
L = \underset{(x,y) \in D}{sup} \Big\|\frac{\partial f(x,y)}{\partial y}\Big\|
\end{equation} 

\subsubsection{Continuous Generative Time Series Models}
We experimented with two architectures employing neural-odes for modeling time series data i.e, ODE-RNN architecture proposed by Rubanova et.al \cite{Ruba2019} \footnote[2]{https://github.com/YuliaRubanova/latent\_ode} and Latent-ODE architecture proposed by \cite{Chen2018} and empirically evaluated the effects of using CCS tuned ODE-Solver such as Nesterov method on their performance. We used the PhysioNet Challenge 2012 dataset \cite{Silva2012} which has ICU Patients conditions observed at different times as time series.
Both of these architectures can be trained as a variational autoencoder to model generative process over time series and are able to handle to non-uniform observation times in the data which eliminates the need for equally-timed binning of observations. The main difference between these models is in their encoder part such that Rubanova et.al \cite{Ruba2019} used ODE-RNN based recognition network as encoder and Chen et.al \cite{Chen2018} used a simple RNN. In Chen et.al's RNN based encoder, the hidden state in-between different observation times remains constant whereas Rubanova et.al showed that modeling the evolution of hidden states in-between observation times as a dynamical process, in the RNN based encoder, better generalizes the hidden state dynamics and improves performance compared to Chen's et.al as well as other autoregressive models such as standard RNN and exponential decay RNN etc, albeit at the cost of higher computational time because we have to solve an ODE at each observation time in the encoder part as well.  \newline
Each time series is modeled as a latent trajectory determined by an initial latent state $z_{t_{0}}$ and a global set of latent dynamics determined by the observation timestamps. The encoder runs backwards in time and outputs the distribution over initial latent state $q_{\phi}(z_{0} | x_{1}, x_{2},\ldots,x_{N})$. The initial state $z_{0}$ is sampled from this distribution and fed to the decoder which is a neural-ODE. Given the observation times $t_{0}, t_{1},\ldots,t_{N}$ and an initial state $z_{0}$, the ODE-solver produces the latent states $z_{t_{1}}, z_{t_{2}},\ldots,z_{t_{N}}$, at each observation time. Finally, the decoder neural network (e.g an MLP) maps these latent states to outputs $x_{1}, x_{2},\ldots,x_{N}$. 
We can extrapolate or interpolate this trajectory arbitrarily far forward or backward in time.
Pseudo-code of algorithm is outline in \ref{algo3}. 

\begin{algorithm}
	\caption{Training Latent ODE-RNN model for Time Series Modeling}
	\label{algo3}
	\KwIn{Data samples with their timestamps ${(x_{i},t_{i})}_{i=1,\cdots,N},$}
	\KwOut{Learned Model}
	\tcp{Encoder}
	$h_{0} = 0$\\
	\For{ i = 1 \KwTo N}{
		$h^{\prime}_{i}$  = ODESolver$(f_{\theta_{enc}}, h_{i-1}, (t_{i-1},t_{i}))$	\tcp{$(f_{\theta_{enc}})$-a NN based approx of encoder ODE-function}
		$h_{i}$ = RNNCell$(h^{\prime}_{i},x_{i})$
		
		$\mu_{z_{0}}, \sigma_{z_{0}}$ = $NN_{enc}(\{h_{i}\}_{i=0}^{N})$ \tcp{a NN to map encoder states to distribution parameters of initial latent state $z_{0}$ }
		$z_{0} \sim \mathcal{N}(\mu_{z_{0}}, \sigma_{z_{0}})$  \tcp{Sample from distribution over latent state $z_{0}$} \label{NN_enc}
		\tcp{Latent states generation}
		$z_{t_{1}},z_{t_{1}},\cdots,z_{t_{M}}$ = ODE-Solver$(f_{\theta_{lat}}, z_{0}, t_{0},\cdots,t_{M})$ \tcp{$(f_{\theta_{enc}})$-a NN based approx of latent state ODE-function}
		\tcp{Decoder}
		$x_{t_{1}},x_{t_{1}},\cdots,x_{t_{M}}$ = $NN_{dec}(z_{t_{1}},z_{t_{1}},\cdots,z_{t_{M}})$ \label{NN_dec} \tcp{a NN to decoder latent state to output values}
		Optimize the model by maximizing ELBO where \\
		ELBO = 	$\sum_{i=1}^{M} log \ref{NN_dec} + log p(z_{t_{0}}) - log \ref{NN_enc}$ where  $p(z_{t_{0}}) = \mathcal{N}(0,1)$ 
	}
	return Learned Model (i.e optimal weights for RNN, $f_{\theta_{enc}}, NN_{enc}, f_{\theta_{lat}}, NN_{dec}$)
	
\end{algorithm}

\begin{table*}[htbp]
	\centering
	\begin{tabular}{|c|c|c|c|}
		\hline
		Method & ODE-Solver & Extrapolation MSE(time in sec) & Interpolation MSE(time in sec) \\
		\hline
		1  & dopri5 & \textbf {0.0045(658)} & \textbf{0.0127(876)} \\
		& nesterov & 0.0061(737) & 0.0268(878) \\
		\hline
		2  & dopri5 & \textbf{0.0047(710)} & \textbf{0.0109}(1025) \\
		& nesterov & 0.0048(778) & 0.0225(\textbf{1020}) \\
		\hline	
		3  & dopri5 & \textbf{0.0044(991)} & \textbf{0.0115(1416)} \\
		& nesterov & 0.0066(1007) & 0.0470(1434) \\
		\hline	
	\end{tabular}
	\caption{Mean square error and training time in sec for extrapolation and interpolation tasks on Physionet time series data. \textbf{Lower is better}. Methods 1,2 and 3 are Latent-ODE(with RNN encoder) \cite{Chen2018}, Latent-ODE(with ODE-RNN encoder) \cite{Ruba2019} and Latent-ODE(ODE-RNN enc + Poisson process modeling of irregular observation times)\cite{Ruba2019} respectively. }
	\label{timeseries}
\end{table*}
\noindent
\cite{Ruba2019} used dopri5 as their default ODE-Solver. We compared Nesterov ODE-Solver with dopri5 for extrapolation and interpolation tasks on time series. Results in Table \ref{timeseries} show that dopri5 outperformed Nesterov by a close margin in both tasks in terms of speed and performance. This is in sharp contrast to its achievement in supervised learning and density estimation experiments. The exact reason for this degradation is not known and needs further exploration. 

\subsubsection{Density Estimation with Continuous Normalizing Flows}
In the third experiment, we used neural-ODE as continuous normalizing flow model for unsupervised density estimation. We used a fast scalable variant of CNF called FFJORD \cite{Grath2019}\footnote[3]{https://github.com/rtqichen/ffjord} to fit the MINIBOONE tabular dataset and MNIST image dataset. Training details are discussed in appendix and pseudo-code of the algorithm is outline in \ref{algo4}.  

\begin{algorithm}
	
	\caption{Density Estimation using FFJORD model}
	\label{algo4}	
	{\LinesNotNumbered
		\KwIn{Dynamics $f_{\theta}$ modeled using a NN, start time $t_{0}$, stop time $t_{1}$, Training data samples x}
		\KwOut{Learned weights $\theta$}
		
		\textbf{function} \hskip 1em $f_{aug}([z_{t} , log p_{t}], t)$: \tcp{Augment f with log-density dynamics}
		\hskip 2em $\epsilon \leftarrow \mathcal{N}(\mu_{1 \times d},\Sigma_{d \times d})$ \tcp{d is the dimensionality of data samples x}
		\hskip 2em $f_{t} = f_{\theta}(z(t),t)$ \tcp{Evaluate neural network $f_{\theta}$}
		\hskip 2em $g \leftarrow \epsilon^{T} \frac{\partial f}{\partial z}\Big|_{z(t)}$ \tcp{Compute vector-Jacobian product}
		\hskip 2em $\tilde{T} = g\epsilon$ \tcp{Unbiased estimate of divergence of dynamics}
		\hskip 2em \textbf{return} $[f_{t}, -\tilde{Tr}]$ \tcp{Concatenate dynamics of state and log-density}
		
	}
	\For{each Batch in Training Data}{
		\For{each sample in Batch}{
			$[z_{0}, \Delta_{log p}] \leftarrow$ odeint$(f_{aug},[x,\vec{0}], t_{0}, t_{1})$\tcp{Solve the CNF ODE}
			log $\hat{p}(x)  \leftarrow $log $p_{z_{0}}(z_{0}) - \Delta_{log p}$ \tcp{Add change in log-density}
			bits-per-dim += -(log $\hat{p}(x)$ - log(256))/log(2) \tcp{compute loss in bits per dimension}
		}
		loss = bits-per-dim/Batch-size \tcp{Averaged over Batch} 
		$\frac{\partial L}{\partial \theta}$ = Adam(loss) \\ 	
		$ \theta = \theta - lr * \frac{\partial L}{\partial \theta}$ \tcp{Optimize weights}
	}
	\textbf{return} $\theta$  \tcp{optimal weights for NN $f_{\theta}$}
\end{algorithm}

\begin{table*}[htbp]
	\centering
	\begin{tabular}{|c|c|c|c|c|}
		\hline
		\multirow{2}{12em}{FFJORD with ODE-Solver} & \multicolumn{2}{c}{MNIST} & \multicolumn{2}{|c|}{MiniBoone} \\
		
		& NLL(bits/dim) & Time & NLL(nats) & Time \\
		\hline
		Euler & 1.6655 & 17 min 15s & -1397 & 6min 10s \\
		\hline
		Nesterov & \textbf{1.563} & 17min & \textbf{-2014} & 3min 40s \\
		\hline
		dopri5 & - & - &- &- \\
		\hline
		AdamBashforth & 1.6863 & 21min 41s & -192 & 4min 4s \\
		\hline
		Runge-Kutta4 & 1.7241 & 22min 31s & -174 & 3min 47s \\
		\hline
	\end{tabular}
	\caption{Negative log-likehood on test data for density estimation task using FFJORD with various ODE-Solvers; \textbf{lower is better}. In nats for Miniboone tabular data and bits/dim for MNIST. - means that training did not complete within the maximum training time set at 1h.}
	\label{density-estimation}
\end{table*}

\noindent
Results in Table \ref{density-estimation} show that Nesterov ODE-Solver outperforms other solvers on both image and tabular datasets. Euler method is the runner-up on both datasets. Surprisingly enough, dopri5 could not even finish training within the set time limit of 1h. Considering the outstanding performance of dopri5 in time series modeling experiment, we observe and hypothesize that the performance of an ODE-Solver is task dependent. 

\section{Conclusion and Future Work}
We presented nesterov gradient descent based ODE-Solver for neural-ode. Our work ensures stability, consistency and faster convergence of training error. This augments current research efforts which mostly focus on faster training through regularization and learning higher-order dynamics. Based on our experiments, we propose following practical takeaways:
\begin{itemize}
	\item We know that not every linear multi-step method obeys CCS conditions. For example, 3-step linear Adams is not zero-stable but a 4-step linear Adam (i.e AdamsBashforth method) is consistent, covergent and zero-stable. So, it is imperative to check CCS (consistency, covergence and zero-stability) conditions of any k-step linear method before using it as ODE-Solver in the Neural-ODE. This is particularly important if you are using a generated or designed ode-solver. The generated coefficients of the solver must satisfy the CCS conditions.
	\item It is possible to achieve significant improvement in speed and performance over ResNet by using a CCS-tuned ODE-solver. Advantage in memory cost has already been established in \cite{Chen2018}.
	\item It is possible that an ODE-solver which has performed remarkably well in a task, fails to do so in some other task. That is true for Nesterov as well other ODE-solvers. Performance is task-dependent. This raises a question: Is there a universal ODE-solver, fit for all tasks? This is an open question and we invite the scientific community to further explore it. 
\end{itemize}
\noindent
Furthermore, there are many other questions to explore. For example, 
\begin{itemize}
	\item Optimization based analogue for an explicit k-step linear method is an open problem. 
	\item \cite{Sci2017} discussed 1-step implicit linear method (i.e Implicit Euler) as an analogue of proximal gradient descent algorithm. Taking inspiration from this result, optimization method based analogues of higher step implicit methods can be explored. Implicit methods although have a higher computational cost because they solve a non-linear system of equation for every new output but are more stable and support lower error tolerance level than explicit methods. 
	\item In nesterov neural-ode, lipschitz constant is used to select the step size. We assumed that true lipschitz constant of gradient flow is known. This could only be possible if you know the ODE you are trying to solve but in most real life cases it is not known and approximated from observations using a surrogate function e.g a neural network. We hypothesize that an accurate estimation of lipschitz constant of neural-ode as fixed step size would further improve results. 
	\item Localized lipschitz constants can make step-size selection in neural ode solvers adaptive, based on the regularity of gradient flow. 
	\item Finally, an incorporation of CCS-tuned ODE-Solver with some regularization approach to smooth the dynamics or learning higher-order dynamics can potentially speed-up the training even more, while ensuring stability and consistency at the same time. 
\end{itemize}

\appendices
	
	\section{Consistency, Covergence and Stability (CCS) conditions}\label{appendix-A}
	We concisely present definitions and results related consistency, stability and convergence of linear multi-step method. Our aim is just to introduce these concepts to the reader without going into technical proofs. Avid readers are encouraged to refer to a standard textbook on the subject for proofs, e.g.,\cite{Suli2003}\cite{Butcher2016}.  \\
	
	\begin{definition}\label{def-stability}\textbf{(Zero-Stability)}
		A linear s-step method is said to be zero-stable if there exists a constant C such that for any two sequences $x_{i}$ and $y_{i}$ that represent two different trajectories of same ode with different initial values, we have
		\begin{multline}\label{sol-diffeq}
		|x_{i} - y_{i}| \le C max\{|x_{0},y_{0}| , |x_{1},y_{1}|, \cdots, |x_{s-1},y_{s-1}|\},\\
		 \hskip 2em \text{as h tends to 0.}
		\end{multline}
	\end{definition}
	
	This equation show that the method is zero-stable if the difference equation \ref{sol-diffeq} has bounded solutions. Zero-stability measures sensitivity of the method to initial conditions; i.e how drastically the solution changes on small perturbations in initial conditions. The algebraic equivalent of zero-stability is known as \textbf{Root Condition}, which we will use to check zero-stability.
	
	\begin{theorem}\textbf{(Root Condition)}\label{root-cond} (see Theorem 12.4 of \cite{Suli2003})
		A linear multi-step method is zero-stable for any initial value problem such as \ref{eq3}, if and only if, all roots of the first characteristics polynomial \ref{lms} of the method are inside the closed unit disc in the complex plane, and any root which lie on the unit circle should be simple.
	\end{theorem}
	
	\begin{definition}\label{consistency}\textbf{(Consistency)}
		A linear multi-step method for an ODE \ref{eq3} is consistent if and only if for any initial condition $x_{0}$, the truncation error (also called the local error) converges to 0 as $h \rightarrow 0$
		\begin{equation}\label{eq-consistency}
		\begin{gathered}
		\lim_{h \rightarrow 0} \norm{T(h)} = 0 , \text{where} \\
		T(h) \triangleq \frac{x(t_{k+s}) - x_{k+s}}{h} 
		\end{gathered}
		\end{equation}
	\end{definition}
	The truncation error $T(h)$ in \ref{eq-consistency} is a measure of error made by the method, normalized by $h$. $x_{k+s}$ is the value obtained by method and $x(t_{k+s})$ is the actual value at time $t_{k+s}$.\\
	We can check consistency in terms of characteristics polynomial using the following proposition:
	\begin{proposition}\label{consistency-coef} (see Proposition 2.4 of \cite{Sci2017})
		A linear multi-step method defined by polynomials $(P,Q)$ is consistent if and only if 
		$$ a(1) = 0 \hspace{5mm} and \hspace{5mm}  \acute{a}(1)= b(1)$$
	\end{proposition}
	
	\begin{theorem}\label{convergence-cond}\textbf{(Covergence)} \textbf{Dahlquist’s Equivalence Theorem:}
		\\For a linear multi-step method $(P,Q)$, consistency and zero-stability are necessary and sufficient conditions for being convergent i.e $x(t_{k}) - x_{k}$ tends to zero for any k when the step size h tends to zero.
	\end{theorem}
	\noindent The Proof is long and technical. See Theorem 6.3.4 of \cite{Walter1997} for details.
	Ensuring that the CCS condition are satisfied requires:
	
	\begin{equation}\label{ccs-cond}
	\begin{aligned}
	a_{2} &= 1 && \text{(Monic polynomial)} \\
	b_{2} &= 0  && \text{(Explicit method)} \\
	a_{0} + a_{1} + a_{2} &= 0   && \text{(Consistency)} \\
	b_{0} + b_{1} + b_{2} &= a_{1} + 2a_{2}   && \text{(Consistency)} \\
	|Roots(P)| &\le 1 && \text{(Zero-stability)}
	\end{aligned}
	\end{equation}

	\section{Datasets and Training}\label{appendix-B}
	We provide additional details about datasets and training here: 
	\subsection*{MNIST Dataset}
	MNIST dataset \cite{lecun2010mnist} consists 28x28 black and white images of numbers from 0 to 9. It has 60k samples for training and 10k for testing. \\
	\noindent 
	\textbf{Supervised Learning:} Out of 60k training samples, we randomly sampled 3k samples for training and 57k for validation. Training epochs were set to 50. Training and testing batch sizes were set to 128 and 1000 respectively. Error tolerance $\mathtt{tol}$ was set to 1e-3. \\
	\noindent
	\textbf{Density Estimation:}
	Out of 60k training samples, we randomly sampled 3k samples for training and 57k for validation. Training and testing batch sizes were set to 200. Model was trained with the Adam optimizer \cite{Kingma2015}. We trained for 1000 epochs with a learning rate of .001 which was decayed to .0001 after 250 epochs.

	\begin{figure}
		\centering		
		\includegraphics[scale = .90]{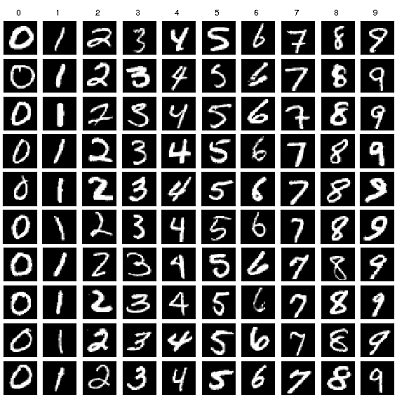}
		\caption{Sample images of MNIST datasets}
	\end{figure}
	\subsection *{PhysioNet Dataset}
	The PhysioNet dataset \cite{Silva2012} consists of observations of 41 features related to patient's condition over a time period of 48 hours. The parameters "Age", "Gender", "Height”, and "ICUType" were removed as these attributes do not vary in time, keeping only 37 features. Measurements for each attribute were quantized by the hour by averaging multiple measurements within the same hour. This reduced the number time stamps, leaving only 49 unique time stamps. We trained the model on this quantized data. The reason for this quantization is to reduce computational cost. In total there are 8000 trajectories. \\
	\noindent 
	\textbf{Time Series Modeling:} We trained on randomly chosen 500 time series samples with batch size 50. The number of latent dimensions of the encoder and decoder were were 40 and 20 respectively. There were 3 encoder and decoder layers and number of units per layer in ODE-Func network and RNN in ODE-RNN recognition network were 50. The number of training epoch were set to 5 and learning rate was set to 1e-2. 
	
	\subsection *{MiniBooNE Dataset}
	This dataset \cite{papamakrios2017} was collected at Fermi-Lab (USA) and has two classes of samples; electron neutrinos (signal) and muon neutrinos (background). Each data sample consists of 43 features. The training set has 29556 samples, the validation set has 3284 samples, and the test set has 3648 samples. \\
	\noindent
	\textbf{Density Estimation:} For the model trained on the MINIBOONE dataset, we used the same architecture as \cite{Grath2019}. The number of epochs was determined adaptively by evaluating the model on the validation set after every 200 iterations and stopping the training once the loss on the validation set did not improve for 30 consecutive epochs. Training and testing batch size was set to 1000. We trained for 1000 epochs with a learning rate of .001 which was decayed to .0001 after 250 epochs. 
	
	\subsection *{Hardware}
	All experiments were run on Tesla T4 GPU with 16 GB RAM.

\bibliographystyle{IEEEtran}
\bibliography{ref}

\begin{IEEEbiography}{Sheikh Waqas Akhtar}{\space} Sheikh Waqas Akhtar received the B.Sc. degree in electrical engineering from the University of Engineering and Technology, Lahore, in 2010 and M.S. degree in computer engineering with the College of Electrical and Mechanical Engineering, National University of Sciences and Technology, Islamabad, in 2017. He is currently serving as Lecturer of Computer Science at University of Central Punjab, Lahore. His research interests include Artificial Intelligence, Machine Learning and Optimization.

\end{IEEEbiography}

\end{document}